\documentclass[conference]{IEEEtran}
\IEEEoverridecommandlockouts

\newcommand{\confbanner}{Work in Process(WiP) Preprint Paper}

\makeatletter
\def\ps@IEEEtitlepagestyle{%
  \def\@oddhead{%
    \parbox[b]{\textwidth}{\centering
      \small\confbanner\par\noindent\rule{\textwidth}{0.4pt}}}%
  \def\@evenhead{}%
  \def\@oddfoot{}%
  \def\@evenfoot{}%
}
\makeatother

\usepackage{cite}
\usepackage[english]{babel}
\usepackage{amsthm}

\usepackage{multirow}
\usepackage{amsmath}
\usepackage{amssymb}
\usepackage{amsfonts}
\usepackage{textcomp}
\usepackage{xcolor}
\usepackage{tikz}
\usepackage{caption}
\usepackage{graphicx}
\usepackage{url}
\usepackage{subfigure}
\usepackage{subcaption}
\usepackage{tabularx}
\usepackage{diagbox}
\usepackage{hyperref}

\def\BibTeX{{\rm B\kern-.05em{\sc i\kern-.025em b}\kern-.08em
    T\kern-.1667em\lower.7ex\hbox{E}\kern-.125emX}}

\renewcommand\thetable{\arabic{table}}
\begin{document}
\newcolumntype{C}[1]{>{\centering\arraybackslash}p{#1}}
\title{Importance Analysis for Dynamic Control of Balancing Parameter in a Simple Knowledge Distillation Setting
}
\author{%
  \IEEEauthorblockN{%
    \parbox{\linewidth}{\centering
      \text{Seongmin Kim},
      \text{ Kwanho Kim$\dagger$},
      \text{ Minseung Kim$\ddagger$},
      \text{Kanghyun Jo*}%
    }%
  }%
  \IEEEauthorblockA{%
    Dept. of Electrical, Electronic and Computer Engineering\\
    Intelligent System Laboratory \\
    University of Ulsan, Ulsan, Korea\\
    \text{\{dailysmile3347, *acejo2208\}@gmail.com},
    \text{\{ $\dagger$ aarony12, $\ddagger$ kmsoiio\}@naver.com}%
  }%
}

\maketitle

\begin{abstract}    
Although deep learning models owe their remarkable success to deep and complex architectures, this very complexity typically comes at the expense of real-time performance.
To address this issue, a variety of model compression techniques have been proposed, among which knowledge distillation (KD) stands out for its strong empirical performance.
The KD contains two concurrent processes: (i) matching the outputs of a large, pre-trained \textit{teacher} network and a lightweight \textit{student} network, and (ii) training the student to solve its designated downstream task.
The associated loss functions are termed the distillation loss and the downsteam-task loss, respectively.
Numerous prior studies report that KD is most effective when the influence of the distillation loss outweighs that of the downstream-task loss.
The influence(or importance) is typically regulated by a balancing parameter.
This paper provides a mathematical rationale showing that in a simple KD setting when the loss is decreasing, the balancing parameter should be dynamically adjusted.
\end{abstract}


\section{Introduction}
Deep learning models have achieved remarkable success in a diverse range of computer vision tasks, such as image classification~\cite{cnn, vit}, object detection~\cite{yolo}, sematic segmentation~\cite{sam}, and image generation~\cite{sde}. 
Underlying these achievements lies an architecture with a deep and complex stack of layers. 
Each layer is made up of composite functions constructed by linear operations (e.g., affine transformations or convolutions) and nonlinear functions known as activation.
Despite their impressive performance, deep learning models with such architectures suffer from limited interpretability and high computational cost.
These shortcomings hinder real-time performance in Internet of Things (IoT), healthcare, and security applications.
Knowledge distillation (KD)~\cite{kd} can alleviate this burden by training a lightweight \textit{student} network under the supervision of a large, pre-trained \textit{teacher} network.
In a standard KD setting, the student network simultaneously minimizes (i) a downstream-task loss (e.g., classification loss) and (ii) a distillation loss that encourages it to replicate the teacher’s results.
Extensive prior research have demonstrated the effectiveness of knowledge distillation-based (KD-based) model compression~\cite{survey}.
Both empirical evidence and theoretical insights indicate that successful KD requires the distillation loss to outweigh the downstream-task loss~\cite{prev1, prev2, prev3}.
However, recent study~\cite{prev4} has shown that relying soley on the distillation loss can adversely affect performance, and some work~\cite{prev5} has proposed methods for determining an optimal balancing parameter between the distillation loss and the downstream-task loss.
To corroborate prior studies, this paper presents a theoretical analysis demonstrating that the balancing parameter should be dynamic rather than a fixed constant.

\section{Theoretical Analysis}
\begin{figure}[htp!]
\centering
\includegraphics[width=9cm, keepaspectratio]{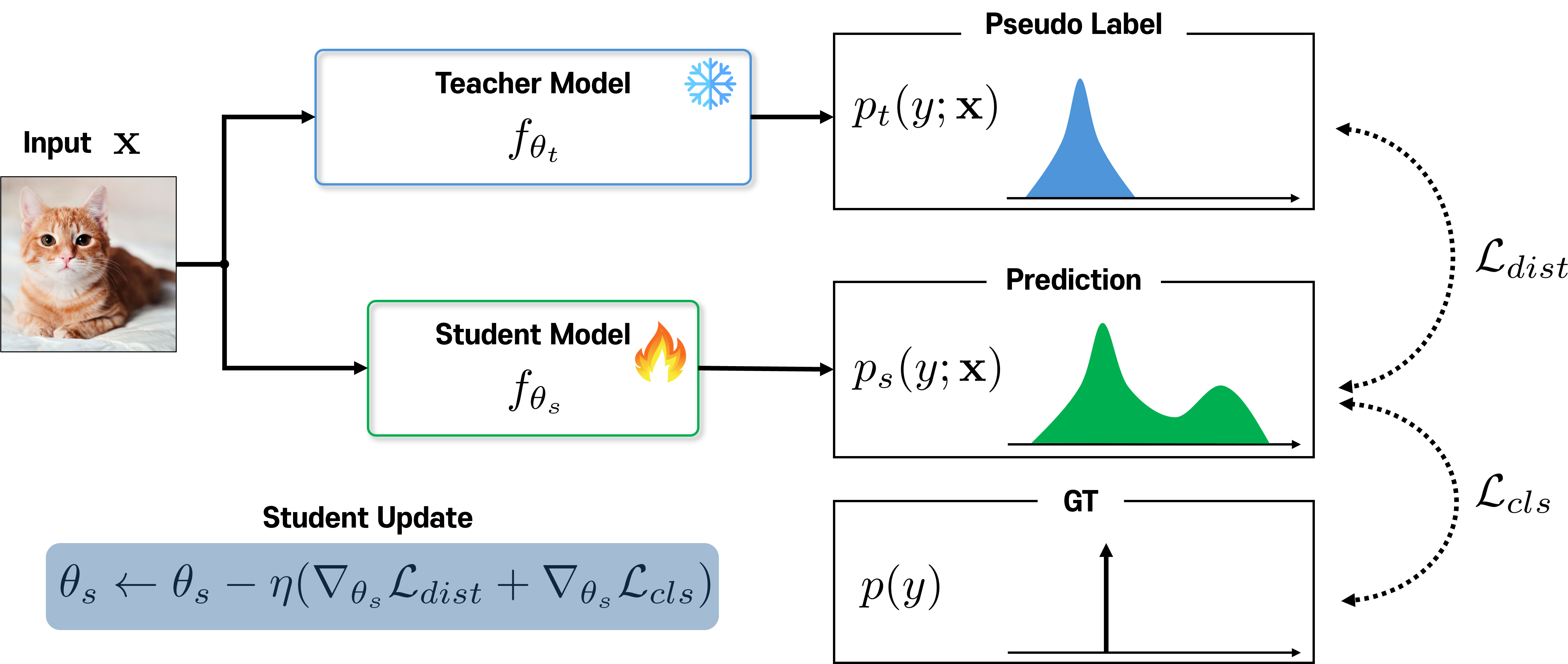}
\caption{This figure depicts the simple knowledge distillation setting of this paper.}
\label{fig1}
\end{figure}

This section presents a mathematical derivation showing that the loss decreases when the model transitions from training step $i$ to step $i+1$.
It then explains how the balancing parameter influences this loss reduction.
This paper assumes an image-classification downstream task.
The detailed situation is illustrated in Fig.~\ref{fig1}.

\subsection{Notations}
The \textit{teacher} model and \textit{student} model are denoted by $f_{\theta_t}$ and $f_{\theta_s}$, respectively, where $\theta_t$ and $\theta_s$ represent their parameters.
Let $\mathcal{L}_{dist}$ and $\mathcal{L}_{cls}$ denote the distillation and classification losses, respectively.
The distillation loss measures the discrepancy between the teacher's and student's class distribution, whereas the classification loss measures the difference between student's prediction and the ground- truth label.
The total loss at $i$th training step is defined as
\begin{equation}
    \mathcal{L}(\theta_s^i) = \lambda \mathcal{L}_{dist}(\theta_s^i) + (1-\lambda)\mathcal{L}_{cls}(\theta_s^i).
\end{equation}
The corresponding gradients are
\begin{equation}
    g^{i}_{dist} = \lambda\nabla_{\theta_s^i} \mathcal{L}_{dist}(\theta_s^i),~~~g^{i}_{cls} =  (1-\lambda)\nabla_{\theta_s^i} \mathcal{L}_{cls}(\theta_s^i).
\end{equation}
Here, $\lambda \in (0, 1)$ denotes the balancing parameter between $\mathcal{L}_{dist}$ and $\mathcal{L}_{cls}$.

\subsection{Loss Difference Analysis}
\begin{figure}[htp!]
\centering
\includegraphics[width=6cm, keepaspectratio]{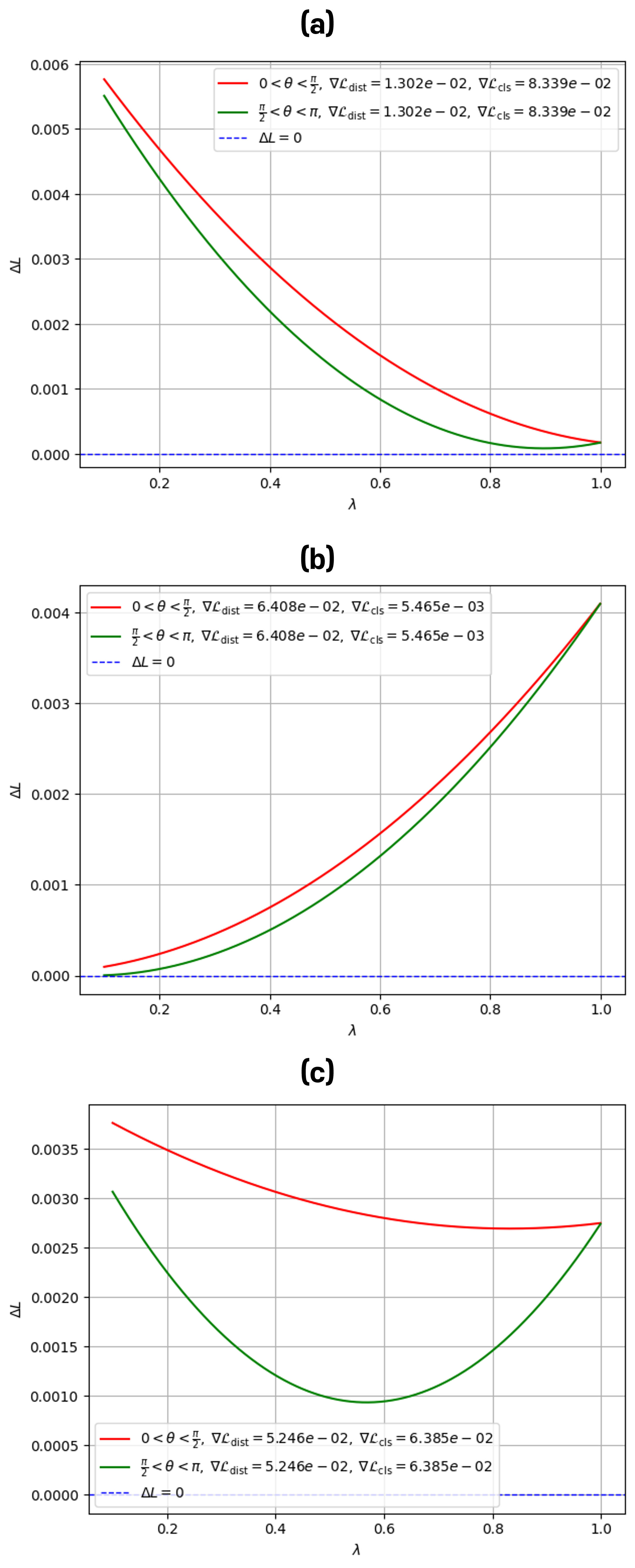}
\caption{The magnitudes of $\nabla_{\theta_s^{i}}\mathcal{L}_{\text{dist}}$ and $\nabla_{\theta_s^{i}}\mathcal{L}_{\text{cls}}$ were independently sampled from a uniform distribution over $[10^{-5}, 10^{-1}]$. The red curve depicts cases in which the angle between the two gradient vectors is acute, whereas the green curve corresponds to an obtuse angle. For each trial, that angle was also sampled uniformly at random within the appropriate range.}
\label{fig2}
\end{figure}

The loss difference at $i+1$ step can be defined as
\begin{equation}
    \Delta \mathcal{L}^{i+1} = \mathcal{L}(\theta_s^{i+1}) - \mathcal{L}(\theta_s^i).
\end{equation}
Here, the total loss at $i+1$ step $\mathcal{L}^{i+1}$ can be represented as
\begin{equation}
    \mathcal{L}(\theta_s^{i+1}) = \mathcal{L} (\theta_s ^i - \eta (g^i_{dist}+g^i_{cls})),
\label{eqn4}
\end{equation}
where $\eta$ denotes the learning rate used in stochastic gradient descent.
Following Fort et al. (2019)~\cite{stiffness}, Equation~\ref{eqn4} can be decomposed via a Taylor expansion.
\begin{align}
\mathcal{L}(\theta_s^{i+1})
&= \mathcal{L}(\theta_s^{i})
   - \eta\,(g_{dist}^{i}+g_{cls}^{i})\cdot\nabla_{\theta_s^{i}}\mathcal{L}
   + \mathcal{O}(\eta^{2}) \\
&\approx \mathcal{L}(\theta_s^{i})
   - \eta\,(g_{dist}^{i}+g_{cls}^{i})\cdot\nabla_{\theta_s^{i}}\mathcal{L}
\end{align}
Then, the loss difference can be rewritten using a first-order Taylor expansion, as shown in the following equations.
\begin{align}
\Delta\mathcal{L}^{i+1}
  &\approx -\eta\bigl(g^{i}_{\text{dist}}+g^{i}_{\text{cls}}\bigr)
           \cdot\nabla_{\theta_s^{i}}\mathcal{L} \\[2pt]
  &=-\eta\!\bigl(
      (g^{i}_{\text{dist}})^{\mathsf T}\nabla_{\theta_s^{i}}\mathcal{L}
      +(g^{i}_{\text{cls}})^{\mathsf T}\nabla_{\theta_s^{i}}\mathcal{L}
    \bigr) \\[2pt]
  &=-\eta\!\bigl(
      (g^{i}_{\text{dist}})^{\mathsf T}g^{i}_{\text{dist}}
      +2(g^{i}_{\text{dist}})^{\mathsf T}g^{i}_{\text{cls}}
      +(g^{i}_{\text{cls}})^{\mathsf T}g^{i}_{\text{cls}}
    \bigr) \\[2pt]
  &=-\eta\Bigl[
      \lambda^{2}\|\nabla_{\theta_s^{i}}\mathcal{L}_{\text{dist}}\|^{2}
      +(1-\lambda)^{2}\|\nabla_{\theta_s^{i}}\mathcal{L}_{\text{cls}}\|^{2} \\
  &\hphantom{=-\eta\Bigl[}\;
      +\,2\lambda(1-\lambda)
        \|\nabla_{\theta_s^{i}}\mathcal{L}_{\text{dist}}\|\,
        \|\nabla_{\theta_s^{i}}\mathcal{L}_{\text{cls}}\|\cos\phi
    \Bigr] \\[2pt]
  &=-\eta\Bigl[\lambda^{2}(\|\nabla_{\theta_s^{i}}\mathcal{L}_{\text{dist}}\|^{2}+\|\nabla_{\theta_s^{i}}\mathcal{L}_{\text{cls}}\|^{2})  \\
  &\hphantom{=-\eta\Bigl[}\;
  -\,2\lambda\|\nabla_{\theta_s^{i}}\mathcal{L}_{\text{cls}}\|^{2} +\|\nabla_{\theta_s^{i}}\mathcal{L}_{\text{cls}}\|^{2} 
    \\
  &\hphantom{=-\eta\Bigl[}\;
  +\,2\lambda(1-\lambda)
        \|\nabla_{\theta_s^{i}}\mathcal{L}_{\text{dist}}\|\,
        \|\nabla_{\theta_s^{i}}\mathcal{L}_{\text{cls}}\|\cos\phi
    \Bigr]
\end{align}
According to the final expression, the loss difference $\Delta \mathcal{L}^{i+1}$ is negative only when the quantity inside the square brackets(i.e., Equation~12) is strictly positive.
Because the term in square brackets equals the squared sum of the two gradients, it is always non-negative.
However, as illustrated in Fig.~\ref{fig2}, the magnitude of the loss reduction varies with the balancing parameter $\lambda$
This occurs because the quantity inside the square brackets is a quadratic function of the balancing parameter $\lambda$.

\section{Discussion}
These results indicate that, at each training step, the influence of the balancing parameter $\lambda$ varies with the angle between the two gradient vectors and their respective magnitudes. Therefore, rather than assigning a fixed value to $\lambda$ throughout training, more effective knowledge distillation requires dynamic adjustment of $\lambda$ at each step in response to current gradients and the desired learning behavior. Here, in the case where the loss should decrease rapidly or converge more gradually, the balancing parameter should vary accordingly. 

\section{Conclusion}
As existing studies show the importance of fine-tuning the balancing parameter in knowledge distillation, this paper offers a mathematical demonstration. When the loss is decreasing, control of $\lambda$ remains necessary, depending on the training state, i.e. the angle and magnitude of the gradients and the required degree of loss reduction.
Future work focuses and investigates an algorithm that will show the results while dynamically adjusting the balancing parameter according to the training state.

\bibliographystyle{unsrt}



\vspace{12pt}

\end{document}